\documentclass[11pt,letterpaper]{article}
\usepackage{cogsys}
\usepackage[T1]{fontenc}
\usepackage{times}
\usepackage[pdftex]{graphicx} 
\usepackage{amsmath}
\usepackage{natbib}
\usepackage{booktabs}
\cogsysheading{X}{20XX}{1-6}{X/20XX}{X/20XX}

\ShortHeadings{Formatting Instructions}
              {Y.\ Xu,  and H.\ Mu\~noz-Avila}

\newcommand{\ChatHTNplus}{\textsc{ChatHTN}$^{+}$}

\usepackage{algorithm}
\usepackage{algpseudocode}

\begin{document} 

\title{Curriculum-Guided Acquisition of Hierarchical Task Knowledge from LLM-Generated Experience}

\author{Yuesheng Xu}{yux220@lehigh.edu}
\author{H\'ector Mu\~noz-Avila}{hem4@lehigh.edu}
\address{Computer Science \& Engineering,
Lehigh University;
Bethlehem, PA  18015-3084  USA}

\vskip 0.2in
 \begin{abstract}
Hierarchical task knowledge enables efficient problem solving by allowing
previously acquired skills to be reused across tasks and problem instances.
In this paper, we investigate curriculum-guided acquisition of hierarchical
task knowledge in an HTN learning framework. We compare three approaches:
online learning during problem solving, human-designed curricula, and an
automated curriculum generation procedure that derives learning experiences
from verified LLM-generated solution traces. The proposed procedure solves
small training problems with an LLM, extracts task accomplishment
information, and transforms it into ordered sequences of state-task learning
problems for HTN method acquisition. We evaluate the three approaches across
multiple planning domains. The results demonstrate that automatically
generated curricula recover much of the benefit of expert-designed curricula
while substantially reducing the token cost of online hierarchical knowledge
acquisition.
\end{abstract}

\section{Introduction}

Hierarchical knowledge plays a central role in intelligent behavior.
Complex tasks are often solved by decomposing them into simpler subtasks,
allowing previously acquired skills to be reused in new situations.
This idea has long been studied in cognitive architectures
\cite{laird2019soar,langley2006unified}, hierarchical reinforcement learning
\cite{barto2003recent}, and Hierarchical Task Network (HTN) planning
\cite{erol1994htn}. By organizing knowledge at multiple levels of abstraction,
hierarchical representations support scalability, transfer, and the acquisition
of increasingly complex skills through composition \cite{langley2006learning}.

The acquisition of such hierarchical structures is a longstanding problem in
cognitive systems. An intelligent agent should not only solve individual
problems, but also acquire reusable procedural knowledge that supports
increasingly complex behavior. In this view, learning a hierarchy involves
constructing higher-level skills from simpler problem-solving experiences and
reusing previously acquired knowledge in new situations. Thus, hierarchical
knowledge acquisition provides a mechanism for the cumulative development of
increasingly sophisticated problem-solving capabilities.

An important question concerns how hierarchical knowledge should be
acquired. Human learners rarely encounter experiences in arbitrary order.
Instead, instruction is often organized around prerequisite skills, with
simpler concepts introduced before more complex ones. This observation has
motivated research on curriculum learning, which studies how the ordering of
learning experiences affects learning efficiency and performance
\cite{bengio2009curriculum}. While curriculum learning has been studied
extensively in machine learning, considerably less attention has been devoted
to how LLM-generated experience can be organized to support the acquisition of
reusable hierarchical procedural knowledge.

Recent advances in large language models (LLMs) create new opportunities for
studying this question. LLMs can generate solutions to planning problems,
decompose tasks into subtasks, and provide explanations of intermediate
reasoning steps. These capabilities suggest that LLM-generated experience can be used to
automatically construct curricula, reducing the need for manually designed
instructional sequences. Because the procedure relies only on task semantics and LLM-generated solution
traces, it can be applied across planning domains without manually designing
domain-specific curricula.

In this paper, we propose an automated curriculum generation procedure for
acquiring hierarchical task knowledge in an HTN learning framework. The
procedure derives ordered learning experiences that support the incremental
acquisition of reusable hierarchical task knowledge from verified
LLM-generated solution traces. It does so by identifying where tasks are
accomplished within successful solution traces and transforming them into
ordered sequences of state--task learning problems. We compare this approach
against two alternatives: online learning during problem solving and
human-designed curricula that sequence learning experiences according to
expert knowledge of the domain.

We evaluate these approaches across multiple planning domains using two
complementary measures: the amount of hierarchical knowledge acquired,
measured by the number of learned methods, and the cost of acquiring that
knowledge, measured through LLM token consumption. The results show that
automatically generated curricula recover much of the benefit of
expert-designed curricula while substantially reducing the cost of online
hierarchical knowledge acquisition.

\section{Illustrative Example}
\label{sec:example}

To illustrate the proposed curriculum generation procedure, consider a problem in the logistics 
transportation domain \cite{} in which a package must be moved from a location in one
city to a destination in another city. The domain contains three hierarchical
tasks. The task \texttt{truckTransport} moves a package within a city,
\texttt{airplaneTransport} moves a package between airports in different cities,
and the top-level task \texttt{transferPackage} combines these capabilities to
deliver a package to its final destination.

Suppose the system generates a small training problem consisting of an initial
state $s_0$ and the top-level task

\[
\texttt{transferPackage(pkg1,locA,locB)}.
\]

\begin{figure}[t]
\centering
\includegraphics[width=\columnwidth]{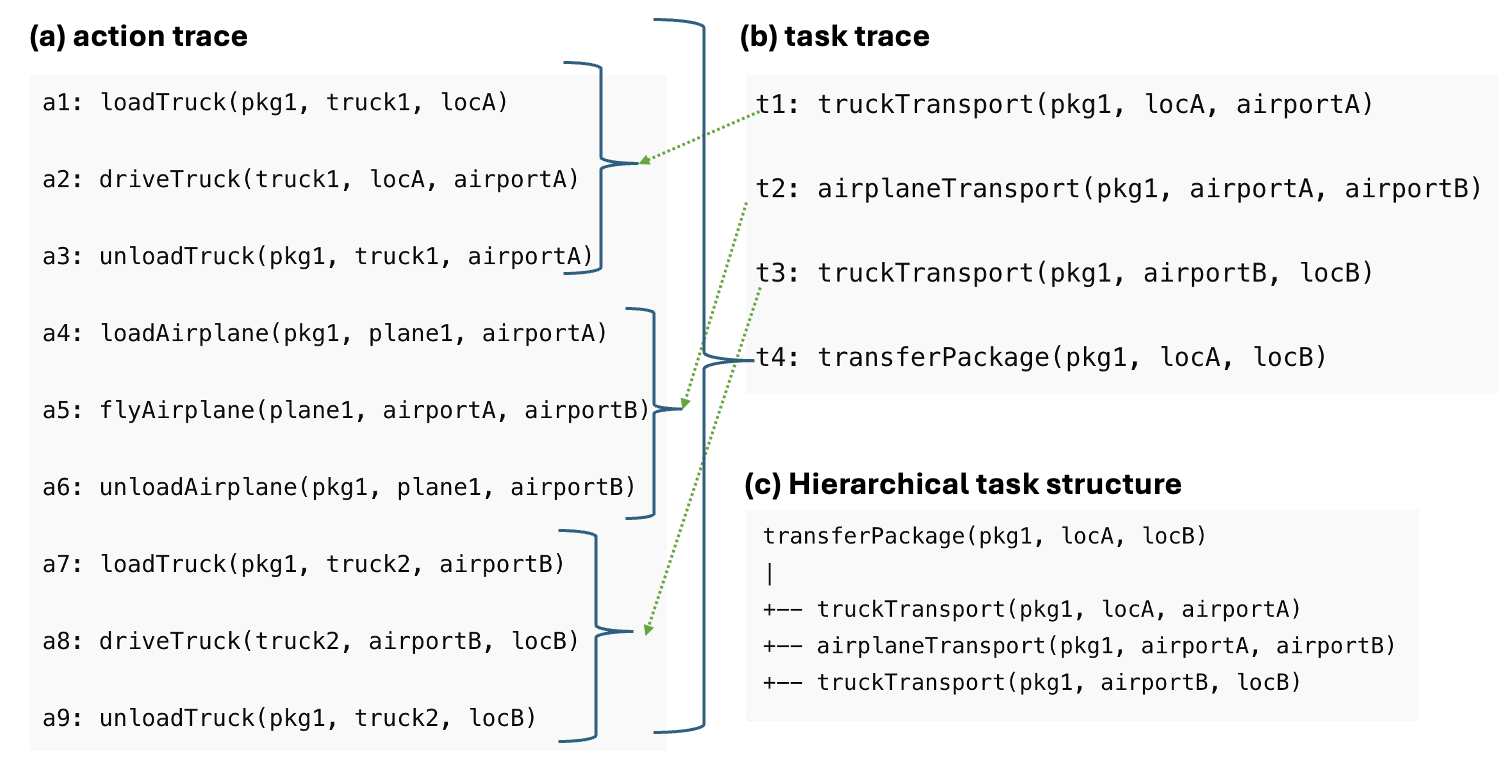}
\caption{
Illustration of the automated curriculum generation procedure.
(a) Action trace generated by the LLM for a small transportation problem.
(b) Corresponding task trace and \textit{(task, action-subsequence)} mappings
indicating the contiguous subsequence of actions over which each task is
executed.
(c) Hierarchical task structure of the top-level task.
}
\label{fig:example}
\end{figure}

Given the primitive operators and task descriptions, the LLM generates a
solution trace. Figure~\ref{fig:example} shows an example. Panel (a) presents
the generated action trace, a plan solving the problem. Panel (b) shows the corresponding task trace,
together with mappings indicating the action at which each task is
accomplished. For example, the first task,

\[
t_1=\texttt{truckTransport(pkg1,locA,airportA)},
\]

is completed by action $a_3$, while the second task,

\[
t_2=\texttt{airplaneTransport(pkg1,airportA,airportB)},
\]

is completed by action $a_6$. Panel (c) depicts the hierarchical task structure
for the problem.

The task-to-action mappings are used to construct a curriculum consisting of
state-task pairs. Let $s_i$ denote the state obtained after executing actions
$a_1,\ldots,a_i$. The first curriculum problem is

\[
(s_0,\texttt{truckTransport(pkg1,locA,airportA)}).
\]

Since this task is completed by action $a_3$, the resulting state is $s_3$.
The second curriculum problem is therefore

\[
(s_3,\texttt{airplaneTransport(pkg1,airportA,airportB)}).
\]

Similarly, because the second task is completed by action $a_6$, the next
curriculum problem is

\[
(s_6,\texttt{truckTransport(pkg1,airportB,locB)}).
\]

The final curriculum problem is

\[
(s_0,\texttt{transferPackage(pkg1,locA,locB)}).
\]

The resulting curriculum introduces tasks in the order in which they are
required by a successful solution. The HTN learning algorithm is trained on these generated problems
sequentially, allowing prerequisite methods to be acquired before
attempting more complex tasks.

\section{Hierarchical Task Network Planning}

Hierarchical Task Network (HTN) planning represents problem-solving knowledge
using tasks, operators, and methods \cite{erol1994htn}. Tasks correspond to
activities that the system seeks to accomplish. Primitive tasks are implemented
directly by operators, whereas compound tasks must be decomposed into simpler
subtasks through methods.

An operator specifies how a primitive task changes the state of the world.
For example, in the Transportation domain, the operator

\begin{quote}
\texttt{loadTruck(pkg1, truck1, locA)}
\end{quote}

loads a package into a truck at a location. Operators are associated with
preconditions that must hold before execution and effects that describe how the
state changes after execution.

Methods specify how compound tasks can be achieved. For example, the task

\begin{quote}
\texttt{transferPackage(pkg1, locA, locB)}
\end{quote}

may be decomposed into the sequence

\begin{quote}
\texttt{truckTransport(pkg1, locA, airportA)}\\
\texttt{airplaneTransport(pkg1, airportA, airportB)}\\
\texttt{truckTransport(pkg1, airportB, locB)}
\end{quote}

which first transports the package to an airport, then flies it between cities,
and finally delivers it to its destination.

Planning proceeds by repeatedly selecting tasks. Primitive tasks are executed
through operators, while compound tasks are decomposed through applicable
methods. This process continues until all remaining tasks are primitive and a
complete executable plan is produced. The resulting hierarchy captures reusable
procedural knowledge that can be applied across multiple problem instances.

Tasks are also associated with semantic descriptions.
Following HTN-MAKER \cite{hogg2008htn}, each task is annotated with
preconditions and effects that characterize when the task can be achieved and
the state changes it is intended to produce. These annotations enable symbolic
reasoning over compound tasks, including verification of task accomplishment,
state projection, and method learning.

\section{Learning Hierarchical Knowledge with ChatHTN+}

The curriculum learning experiments in this paper use
\ChatHTNplus\ \cite{xu2025online}, an HTN learning framework that combines symbolic
planning with large language models. The system initially contains the primitive
operators of a domain but no methods. Consequently, compound tasks cannot be
decomposed using domain-specific hierarchical knowledge.

When the planner encounters a compound task for which no applicable method
exists, it queries an LLM to generate a candidate decomposition. Unlike earlier
versions of the framework that restricted decompositions to primitive actions,
the version used in this paper allows generated decompositions to contain both
primitive and compound tasks. This enables the acquisition of hierarchical
abstractions at multiple levels of granularity.

Generated decompositions are validated using the semantic descriptions of tasks
and operators. If a decomposition can be successfully incorporated into a
solution, the system learns a new HTN method that captures the decomposition and
adds it to its knowledge base. Learned methods are subsequently reused when
solving future problems.

As planning and learning proceed, the method set grows incrementally, allowing
the system to acquire increasingly complex hierarchical knowledge. The central
question considered in this paper is how different curriculum-generation
strategies influence the efficiency with which this knowledge is acquired.

\section{Automated Curriculum Generation}
\label{sec:curriculum}

\subsection{Overview}

The automated curriculum generation procedure takes as input a set of tasks
$T$, a set of operators $O$, and the desired number of training problems $n$
(line~1 of Algorithm~\ref{alg:curriculum}). Its output is a curriculum $C$
consisting of state-task pairs that can be used to train an HTN learning
algorithm (line~2).

The procedure repeatedly generates small random training problems and queries
an LLM for solution traces. The generated traces are then validated using the
semantic descriptions of tasks and operators. From each verified trace, the
procedure extracts task accomplishment information and constructs a sequence of
state-task pairs. These pairs are added to the curriculum and subsequently
used to train the HTN learning algorithm. Figure~\ref{fig:example}
illustrates this process for a transportation problem.

\begin{algorithm}[htb]
\caption{GenerateCurriculum}
\label{alg:curriculum}
\begin{algorithmic}[1]
\State {\bfseries Input:} Tasks $T$, operators $O$, number of training problems $n$
\State {\bfseries Output:} Curriculum $C$
\State $C \leftarrow ()$
\For{$i = 1$ to $n$}
    \State Generate random small problem $(s_0,t)$ with $t \in T$
    \State $(\tilde{a},\tilde{t},\mu) \leftarrow \textsc{LLM}(s_0,t,T,O)$
    \If{\textsc{Verify}$(\tilde{a},O,s_0,t)$}
        \State $\tilde{s} \leftarrow \textsc{Project}(s_0,\tilde{a},O)$
        \ForAll{$(t',a) \in \mu$}
            \State let $s \in \tilde{s}$ be the state immediately after $a$
            \State Add $(s,t')$ to $C$
        \EndFor
        \State Add $(s_0,t)$ to $C$
    \EndIf
\EndFor
\State {\bfseries return} $C$
\end{algorithmic}
\end{algorithm}

\subsection{From Solution Traces to Learning Experiences}

Algorithm~\ref{alg:curriculum} generates a curriculum $C$ from a collection of
randomly generated training problems. The curriculum is initialized to the
empty sequence at line~3. For each iteration (line~4), a random small
problem $(s_0,t)$ is generated (line~5), where $t$ is a top-level task in $T$
and $s_0$ is a randomly generated initial state. The LLM is then queried to 
produce an explicitly indexed trace of primitive actions 
$\tilde{a} = [(1, a_1), \ldots, (k, a_k)]$, an explicitly indexed 
trace of compound tasks in depth-first order 
$\tilde{t} = [(1, t_1), \ldots, (l, t_l)]$, and an interval mapping 
$\mu$ (line~6). This mapping assigns each task index to the contiguous interval 
of action indices that realize it. Specifically, for a task index $i$, 
$\mu(i) = (j_{\text{start}}, j_{\text{end}})$ indicates that task $t_i$ begins 
at action index $j_{\text{start}}$ and is accomplished immediately after 
executing action $a_{j_{\text{end}}}$.

Before a curriculum is generated, the returned action trace is validated using the
operators (line~7). Verification
symbolically executes $\tilde{a}$ from $s_0$ and checks that all operator
applications satisfy their preconditions and effects. The resulting state is then checked against the semantic description of the top-level task $t$ to ensure that the generated solution indeed accomplishes the requested task. Only traces that satisfy these checks are retained for
curriculum generation.

From a single validated plan we then derive one problem/task pair $(s, t')$ for
every compound task $t'$ in the task trace. Each pair is obtained by replaying the
plan once from the original initial state $s_0$ and reading off, from the
task-to-action mapping, the state $s$ that holds immediately before the actions
realizing $t'$ begin. A single top-level problem thus fans
out into a set of problem/task pairs, each posing one compound task to be solved from the state
the top-level plan reached it in. The pairs are
ordered by the level of their task in the decomposition, from the lowest-level
compound tasks (those that decompose directly into primitive sub-tasks) up to the
original top-level task. This sequence defines a curriculum built on cumulative difficulty, ensuring that complex tasks are only attempted once their constituent parts are no longer a barrier. Thus, when a higher-level task is learned, the methods for the subtasks it depends on have already been acquired and stored in memory.

After generating curriculum problems for all mappings, the original training
problem $(s_0,t)$ is added as the final curriculum problem (line~13). This
allows the learner to attempt the top-level task after acquiring experience
for the constituent tasks that appear in its solution. The complete curriculum
is returned at line~16.

The generated curriculum is presented sequentially to the HTN learning
algorithm. Because curriculum problems are generated from task accomplishments along a
successful solution trace, tasks are introduced in the order in which they are
required by that solution. Consequently, the resulting
curriculum captures prerequisite relationships implicit in the solution trace
without requiring manual curriculum design.

\section{Empirical Evaluation}
\subsection{Domains}
We conduct an empirical evaluation across four planning domains. These include the IPC Factory domain \cite{IPC2020}, two classic planning environments: Transportation and SearchAndRescue, and Mystery. Mystery is a semantically obfuscated domain that mirrors the underlying logic of Transportation but substitutes standard vocabulary with random lexical equivalents. Mystery evaluates whether the proposed approach can acquire and exploit hierarchical task knowledge when semantic cues from domain and predicate names are unavailable, forcing the LLM to rely on the provided task and operator descriptions rather than familiar planning terminology.

\subsection{ Setup}
We evaluate three configurations that share the same HTN planner (ChatHTN+),
the same underlying domains, and the same test problems, differing only
in whether and how the domain knowledge is acquired before testing. 

The first
configuration is our curriculum approach, which learns a set of
reusable, HTN methods from tiny scale problems. The resulting domain knowledge is evaluated on small,
medium, and large testing problems. 

The second configuration is a
human-crafted baseline: rather than
generating problems and querying LLM to solve them automatically, a human expert supplies a
compact set of problem/task pairs that is already sufficient to induce the
complete domain knowledge. The system learns from this curated set and is then tested on
the identical small, medium, and large problems. 

The third configuration is standard  \ChatHTNplus\
 with
empty-memory that begins with no methods at all.  \ChatHTNplus\ is run directly on the small, medium, and large test
problems and is forced to learn methods on demand, decomposing each problem from
the top-level task downward whenever no applicable method exists in memory. Consequently, knowledge acquisition and
problem solving occur simultaneously throughout testing.

We report the comparison of
these three configurations across all four domains.

The system parameter for the prompt limit was configured to a maximum of 30. Table~\ref{tab:state-axioms} reports the average number of state axioms per problem instance for each domain and problem-size category (curriculum, small, medium, large).

\begin{table}[t]
  \centering
  \caption{Average number of state axioms per problem, by domain and problem size.}
  \label{tab:state-axioms}
  \begin{tabular}{lrrrr}
    \toprule
    Domain & Curriculum & Small & Medium & Large \\
    \midrule
    Factory         & 54 & 442 & 1042 & 2082 \\
    Mystery         & 54 & 406 & 806  & 1606 \\
    SearchAndRescue             & 32 & 308 & 408  & 608  \\
    Transportation  & 54 & 406 & 806  & 1606 \\
    \bottomrule
  \end{tabular}
\end{table}

\subsection{Empirical Results}

\begin{figure}[htbp]
    \centering
    \includegraphics[width=\columnwidth]{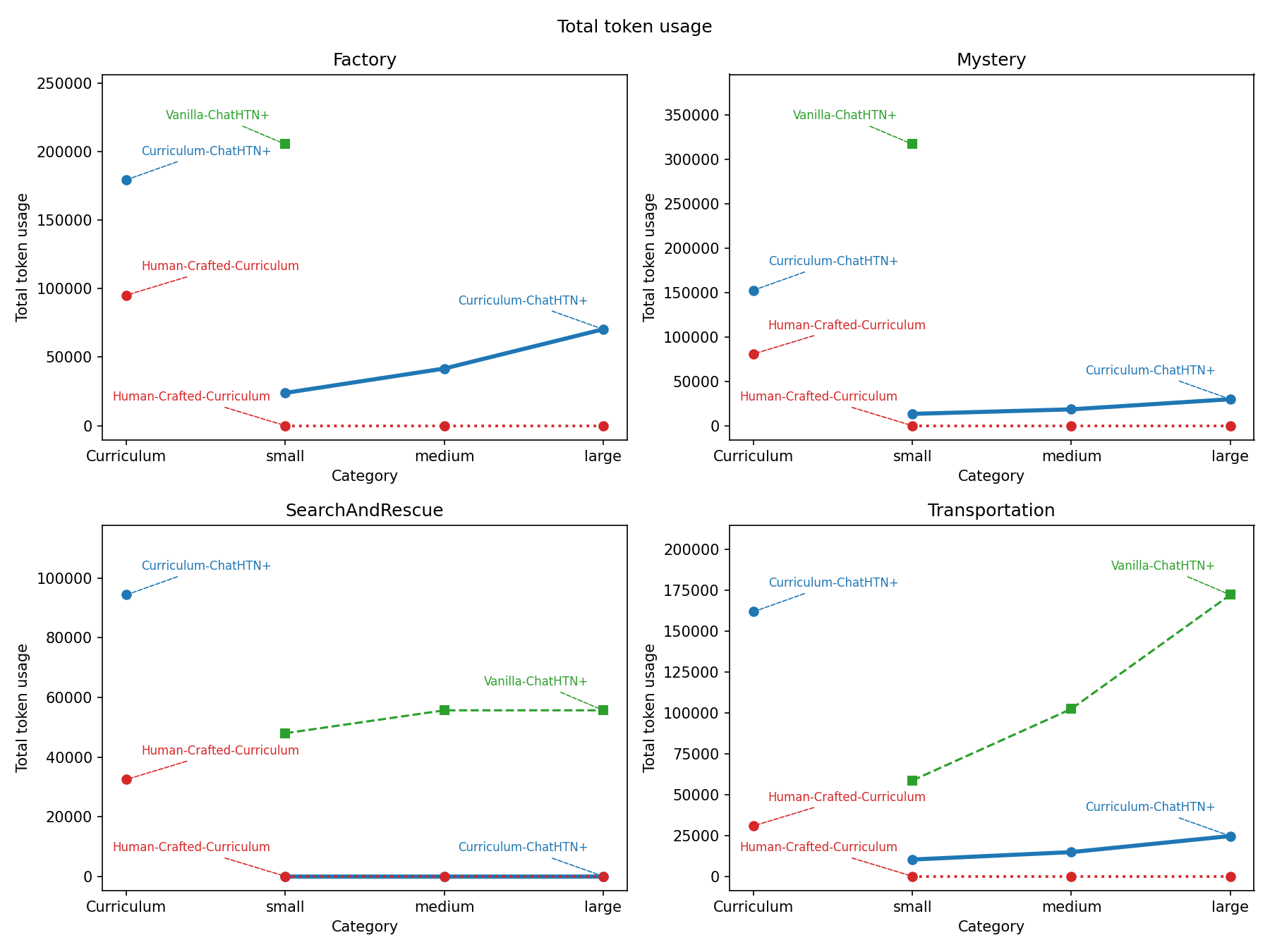}
    \caption{Total token usage across the Factory, Mystery, SearchAndRescue, and Transportation domains. Curriculum-ChatHTN+ incurs an initial overhead during the curriculum learning stage, but this structured approach is essential for acquiring complete domain knowledge. Without it, Vanilla-ChatHTN+ fails on medium and large problems in complex situations. The human-crafted baseline serves as the theoretical upper bound for curriculum learning.}
    \label{fig:figure2}
\end{figure}

Figure~\ref{fig:figure2} reports the total token consumption for the three
learning configurations across the four planning domains. The first category
("Curriculum") corresponds to the cost of acquiring hierarchical task
knowledge before testing. The remaining categories report the token
consumption incurred while solving the test problems.

Curriculum-ChatHTN+ incurs an upfront cost because it acquires reusable HTN methods before testing. Once these methods have been learned, they can be reused across subsequent problem instances, substantially reducing the need for additional LLM queries

Our experimental results indicate that without curriculum learning, the baseline Vanilla-ChatHTN+ is only capable of solving relatively simple domains, such as Transportation and SearchAndRescue. The primary limitation of this system lies in its heavy reliance on the LLM at every task decomposition occurring during planning. Specifically, at any given state for a specific task, the queried LLM must generate a decomposition that is accurate not only locally, but also globally. Although the LLM typically generates locally appropriate decompositions to resolve these localized task decompositions when queried with small-scale subproblems. However, this localized querying strategy introduces a critical vulnerability: the LLM will occasionally propose decompositions that are locally valid but globally flawed.

If the planner traverses an incorrect path and reaches a dead-end state, the system is forced to backtrack. Because the exact point of failure is unknown, the system must exhaustively backtrack step-by-step, necessitating repeated LLM queries until the initial mistake is identified. As the problem scale increases, these missteps lead the planner deeper into extended erroneous paths, making recovery increasingly difficult and resource-intensive. Because these incorrect paths can be arbitrarily long, Vanilla-ChatHTN+ failed on medium and large problem sizes in the Factory and Mystery domains after exceeding the maximum allowable prompt threshold. Consequently, for complex domains, Vanilla-ChatHTN+ struggles to solve the entire domain starting strictly from a top-level task.

In contrast, our proposed method, Curriculum-ChatHTN+, successfully learns the vast majority of necessary HTN methods during its training phase. Any remaining unrepresented edge cases can be effectively handled by the online learning mechanism of ChatHTN+ when specific problems are encountered during testing. Expectedly, Curriculum-ChatHTN+ incurs a substantial initial overhead during the curriculum learning stage, which is inevitable when aiming to acquire a relatively complete domain model from scratch in a general domain.

In the two simpler domains, Curriculum-ChatHTN+ consumes a higher total number of tokens because the system must pay this initial overhead—specifically, utilizing the LLM as a planner to solve minimal-scale problems, which dominates a significant portion of the overall cost. However, in the two complex domains, Curriculum-ChatHTN+ utilizes a comparable or even lower number of tokens on small-scale test problems. This improved efficiency arises because the token cost required to learn a method during the curriculum learning stage is significantly lower than the cumulative cost incurred by online learning and backtracking during the testing phase.

Finally, The human-designed curriculum provides a reference point that approximates the
best achievable curriculum for these domains. It empirically verifies the existence of a specific set of problem/task pairs from which complete domain knowledge can be seamlessly acquired. By leveraging these hand-crafted problem/task pairs, the system extracts all necessary methods prior to testing. As a result, across all domains and problem scales, this approach entirely eliminates the need for further LLM queries during execution. However, achieving this optimal performance comes at the steep cost of requiring extensive domain expertise and significant human effort to manually author the optimal curriculum.

The experimental results demonstrate that automatically generated curricula recover much of the benefit of expert-designed curricula while dramatically reducing the cost of online hierarchical knowledge acquisition.

Beyond the number and acquisition cost of the learned methods, the results also provide evidence of their utility as reusable procedural knowledge. Curriculum-ChatHTN+ acquires its methods from tiny training problems and
subsequently reuses them to solve independently generated test problems at small, medium, and large scales. Thus, the acquired knowledge is not specific to the training instances from which it was learned. In particular, successful
reuse on larger test problems demonstrates that the learned methods generalize across problem instances and problem scales. Moreover, the reduced need for additional LLM queries during testing shows that this knowledge can be applied
directly to new problems rather than requiring substantial relearning.

\section{Related Work}
\subsection{HTN Learning}

Research on the acquisition of hierarchical planning knowledge has a long
history in AI and cognitive systems. Early cognitive architectures such as
ICARUS \cite{langley2005interleaving} emphasized the learning and use of
hierarchically organized procedural knowledge, motivated by the observation
that complex tasks can often be solved more efficiently through recursive
decomposition into reusable skills. More generally, researchers have argued
that hierarchical knowledge plays a central role in intelligent behavior
because it reduces search, supports transfer across tasks, and provides
interpretable representations of problem-solving expertise
\cite{langley2025learning}. These ideas strongly influenced subsequent research on hierarchical planning, including HTN planning, where procedural knowledge is represented explicitly as methods.

Within the HTN framework, research has investigated the automatic acquisition
of hierarchical planning knowledge from demonstrations, execution traces,
problem-solving experience, and domain descriptions. HTN-MAKER
\cite{hogg2008htn,hogg2016learning} learns methods by analyzing solution
traces together with semantically annotated tasks. Subsequent work extended
HTN learning to partially observed traces \cite{zhuo2009learning,zhuo2014learning},
case-based learning \cite{ilghami2002camel}, probabilistic HTNs \cite{li2010learning}. Collectively, these approaches seek to reduce the knowledge-engineering burden associated with manually constructing HTN methods by learning them from different forms of experience.

Beyond learning individual HTN methods, several researchers have investigated the automatic discovery of hierarchical task structures themselves. Word2HTN \cite{gopalakrishnan2016word2htn} combines
distributional semantic representations and goal reasoning to infer task
hierarchies from solution traces, while landmark-based approaches use
planning structure to guide method acquisition
\cite{fine2020learning}. More recently, CURRICULAMA
\cite{li2024automatically} introduced a curriculum-based approach to HTN learning.
CURRICULAMA extracts planning landmarks from classical planning problems and
uses them to generate curricula that progressively introduce increasingly
complex learning problems. The resulting curricula are then used to guide
HTN-MAKER's method-learning process.

In most of these systems, the focus is on the learning algorithm itself,
whereas the organization of learning experiences is either predetermined or
treated as secondary.

Our work is complementary to these approaches. Rather than proposing a new
HTN-learning algorithm, we investigate how the acquisition process itself can
be organized more efficiently. In particular, we focus on automatically
constructing learning curricula from verified LLM-generated solution traces.
Whereas CURRICULAMA derives curricula from symbolic analyses of planning
problems, our approach derives them from successful LLM-generated execution
traces, transforming task accomplishment information into ordered
state--task learning problems that support the acquisition of reusable HTN
methods. Thus, the curriculum is induced directly from problem-solving
experience rather than from symbolic analyses such as landmark extraction.

Although previous work has explored a variety of mechanisms for acquiring hierarchical planning knowledge, comparatively little attention has been devoted to the cost of knowledge acquisition itself. Existing evaluations have typically emphasized planning performance, convergence, or the number of learned methods. In contrast, our work explicitly investigates the efficiency of hierarchical
knowledge acquisition by measuring the LLM token expenditure required to
construct reusable hierarchical knowledge under different
curriculum-generation strategies.

\subsection{LLMs for Planning and Knowledge Acquisition}

Recent advances in large language models have renewed interest in their use for planning, reasoning, and knowledge acquisition. Some approaches employ LLMs directly as planners,
using prompting strategies to generate plans, reason about action sequences,
or decompose problems into intermediate steps. 
Representative examples include Chain-of-Thought prompting
\cite{wei2022cot}, ReAct \cite{yao2022react},
Plan-and-Solve \cite{wang2023plan},
Tree-of-Thoughts \cite{yao2023tree}, and
Reasoning via Planning (RAP) \cite{hao2023reasoning}. These systems demonstrate that LLMs can perform impressive
forms of planning and reasoning, particularly in domains that require substantial commonsense or semantic knowledge.

Other research has combined LLMs with symbolic planning systems. Such
approaches use symbolic representations to constrain, verify, or refine
LLM-generated solutions. Examples include systems that integrate LLMs with
classical planners, task planners, robotic controllers, and agent
architectures. These hybrid approaches seek to combine the flexibility of
LLMs with the reliability and interpretability of symbolic reasoning. More generally, these systems illustrate the complementary strengths of LLMs and symbolic reasoning: LLMs provide flexible knowledge acquisition and generation, whereas symbolic systems provide verification, search, and execution guarantees.

A particularly relevant line of research uses LLMs not merely as planners but
as sources of reusable planning knowledge. Recent work has investigated the automatic generation of action models,
planning operators, task decompositions, procedural knowledge, and HTN
methods from natural language descriptions and problem-solving traces \cite{tantakoun2025llms}. These efforts
suggest that LLMs can serve as knowledge-acquisition mechanisms capable of
producing symbolic structures that can subsequently be validated and reused.

ChatHTN+ \cite{xu2025online} combines HTN planning with LLM-generated task
decompositions and online method learning. By validating generated
decompositions against symbolic task semantics, the framework transforms
verified LLM-generated decompositions into reusable HTN methods that
accumulate over successive planning episodes. However, the system acquires
methods opportunistically, learning them only when they are first required
during search. The present work builds directly on this capability by
investigating how automatically generated curricula can organize these
learning experiences before testing begins. 

Existing work has focused primarily on using LLMs either to solve planning
problems or to generate reusable planning knowledge. In contrast, we treat
verified LLM-generated solution traces as a source of experience from which
learning curricula can be constructed automatically. Rather than learning
directly from isolated decompositions or demonstrations, we transform
LLM-generated solution traces into ordered sequences of state--task learning
problems that support the efficient acquisition of reusable hierarchical task
knowledge.

\subsection{Curriculum Learning}

Curriculum learning studies how the ordering of learning experiences influences
learning efficiency and performance. Inspired by the observation that humans
often acquire skills through structured instruction, Bengio et al.
\cite{bengio2009curriculum} proposed presenting training examples in a
meaningful order, typically progressing from simpler concepts to more complex
ones. Since then, curriculum learning has been investigated extensively in
machine learning, where it has been shown to accelerate convergence, improve
generalization, and facilitate the acquisition of difficult concepts.
This observation is particularly relevant for hierarchical learning settings,
where higher-level skills often depend on the prior acquisition of simpler
procedural knowledge.

A central theme in this literature is the automatic generation of curricula.
Rather than relying on manually designed training sequences, researchers have
proposed methods that derive curricula from measures of task difficulty,
estimates of learner competence, learning progress, or other properties of the learning process \cite{wang2021survey}. Closely related work on self-paced learning
allows a learner to determine the order in which experiences are acquired,
typically favoring examples that are neither too easy nor too difficult
\cite{kumar2010self}.

Curriculum learning has also been studied in reinforcement learning, skill
acquisition, and developmental robotics. In these settings, curricula are often
used to organize the acquisition of increasingly complex behaviors by
introducing prerequisite skills before more complex skills
\cite{narvekar2020curriculum}. Such approaches are motivated by the observation that complex
behavior frequently depends on the successful acquisition of simpler skills,
making the ordering of experiences an important factor in learning efficiency.

Our work differs from traditional curriculum-learning approaches in several
respects. First, unlike most curriculum-learning approaches, which aim to improve the optimization of statistical models, our objective is to improve the efficiency of acquiring reusable symbolic procedural knowledge.
Second, the curriculum is derived from verified LLM-generated experience.
Third, the curriculum is constructed from task accomplishment information
embedded within solution traces, allowing prerequisite relationships among
tasks to be extracted automatically from successful problem-solving episodes.
Finally, rather than evaluating curricula solely in terms of learning
performance, we explicitly study their effect on the efficiency of
hierarchical knowledge acquisition through LLM token consumption.
Consequently, the proposed approach combines ideas from curriculum learning,
hierarchical knowledge acquisition, and LLM-based planning.

\section{Conclusion}

Hierarchical task knowledge enables intelligent systems to accumulate
procedural knowledge that can be reused across increasingly complex problems.
In this paper, we investigated how such knowledge can be acquired efficiently
through curriculum-guided learning. We introduced an automated curriculum
generation procedure that transforms verified LLM-generated solution traces
into ordered sequences of state--task learning problems. These learning
experiences are subsequently used by an HTN learning algorithm to acquire
reusable methods before encountering more challenging problems.

Our empirical evaluation compared automated curricula with both
human-designed curricula and purely online learning. Across four planning
domains, the results showed that automatically generated curricula recover
much of the benefit of expert-designed curricula while substantially reducing
the token cost of online hierarchical knowledge acquisition. In the more
complex domains, curriculum-guided learning also enabled the acquisition of
hierarchical knowledge that could not be obtained reliably through online
learning alone because of the prohibitive cost of repeated LLM queries and
backtracking.

More broadly, the results suggest that LLMs can play a role beyond serving as
problem solvers or decomposition generators. Their generated solutions can be
treated as a source of experience from which reusable symbolic knowledge is
systematically acquired. This perspective opens new opportunities for
combining generative AI with symbolic learning systems, allowing LLM-generated
experience to support the cumulative acquisition of hierarchical procedural
knowledge.

Several directions remain for future research. The current work assumes that
curricula are generated from collections of randomly generated small training
problems. An interesting direction is to generate these problems
adaptively, selecting new training problems based on the learner's current
knowledge and remaining deficiencies. Another direction is to investigate
alternative criteria for constructing curricula from solution traces,
including measures of task dependency, method utility, or expected learning
benefit. Finally, although our experiments focused on HTN learning, the
general idea of deriving curricula from verified LLM-generated experience may
also apply to other forms of symbolic knowledge acquisition.


\begin{acknowledgements} 
\noindent
This material is related to work supported by the National Science Foundation under Grant No. 2628497. Any opinions, findings, and conclusions or recommendations expressed in this material are those of the author(s) and do not necessarily reflect the views of the National Science Foundation.
\end{acknowledgements}

{\parindent -10pt\leftskip 10pt\noindent
\bibliographystyle{cogsysapa}
\bibliography{format}

}


\end{document}